\documentclass[10pt,twocolumn,letterpaper]{article}

\usepackage{cvpr}              

\usepackage{graphicx}
\usepackage{amsmath}
\usepackage{amssymb}
\usepackage{booktabs}
\usepackage{float}
\usepackage{color}
\usepackage{multirow}
\usepackage{indentfirst}
\usepackage[accsupp]{axessibility}

\usepackage[pagebackref,breaklinks,colorlinks]{hyperref}

\usepackage[capitalize]{cleveref}
\crefname{section}{Sec.}{Secs.}
\Crefname{section}{Section}{Sections}
\Crefname{table}{Table}{Tables}
\crefname{table}{Tab.}{Tabs.}

\def\cvprPaperID{4898} 
\def\confName{CVPR}
\def\confYear{2023}

\begin{document}

\title{GarmentTracking: Category-Level Garment Pose Tracking}
\author{Han Xue\textsuperscript{2,1}, Wenqiang Xu\textsuperscript{2}, Jieyi Zhang\textsuperscript{2}, Tutian Tang\textsuperscript{2}, Yutong Li\textsuperscript{2}, Wenxin Du\textsuperscript{2}, Ruolin Ye\textsuperscript{3}, Cewu Lu\textsuperscript{$\dagger$1,2}\\
\textsuperscript{1}Shanghai Qi Zhi Institute
\textsuperscript{2}Shanghai Jiao Tong University
 \textsuperscript{3}	Cornell University\\
\textsuperscript{2}{\tt\small \{xiaoxiaoxh, vinjohn, yi\_eagle, tttang, davidliyutong, mnkmYuki, lucewu\}@sjtu.edu.cn}\\
\textsuperscript{3}{\tt\small ry273@cornell.edu}
}
\maketitle

\begin{abstract}
   Garments are important to humans. A visual system that can estimate and track the complete garment pose can be useful for many downstream tasks and real-world applications. In this work, we present a complete package to address the category-level garment pose tracking task: (1) A recording system \textbf{VR-Garment}, with which users can manipulate virtual garment models in simulation through a VR interface. (2) A large-scale dataset \textbf{VR-Folding}, with complex garment pose configurations in manipulation like flattening and folding. (3) An end-to-end online tracking framework \textbf{GarmentTracking}, which predicts complete garment pose both in canonical space and task space given a point cloud sequence. Extensive experiments demonstrate that the proposed GarmentTracking achieves great performance even when the garment has large non-rigid deformation. It outperforms the baseline approach on both speed and accuracy. We hope our proposed solution can serve as a platform for future research. Codes and datasets are available in \url{https://garment-tracking.robotflow.ai}.
\end{abstract}

\renewcommand{\thefootnote}{}
\footnotetext{$\dagger$ Cewu Lu is the corresponding author, the member of Qing Yuan Research Institute and MoE Key Lab of Artificial Intelligence, AI Institute, Shanghai Jiao Tong University, China and Shanghai Qi Zhi Institute.}

\section{Introduction}
\label{sec:intro}
Garments are one of the most important deformable objects in daily life. A vision system for garment pose estimation and tracking can benefit downstream tasks like MR/AR and robotic manipulation\cite{ha2022flingbot, avigal2022speedfolding}. 
The category-level garment pose estimation task is firstly introduced in GarmentNets \cite{chi2021garmentnets}, which aims to recover the full configuration of an \textbf{unseen} garment from a single static frame.
Unlike the non-rigid tracking methods \cite{newcombe2015dynamicfusion, innmann2016volumedeform, guo2017real, slavcheva2017killingfusion, slavcheva2018sobolevfusion, bozic2020deepdeform, bozic2020neural} which can only recover the geometry of the visible regions, pose estimation task can also reconstruct the occluded parts of the object. Another line of works \cite{dou2016fusion4d, dou2017motion2fusion, li20214dcomplete, niemeyer2019occupancy, palafox2021npms, palafox2022spams, lin2022occlusionfusion} (non-rigid 4D reconstruction which can reconstruct complete object geometry) cannot be directly applied on garments, since they assume the object has a watertight geometry. In contrast, garments have thin structures with holes.

In this paper, we propose a new task called \textbf{Category-level Garment Pose Tracking}, which extends the single-frame pose estimation setting in \cite{chi2021garmentnets} to pose tracking in dynamic videos. Specifically, we focus on the pose tracking problem in garment manipulation (\eg flattening, folding). In this setting, we do not have the priors of the human body like previous works for clothed humans \cite{pons2017clothcap, zhu2020deep, ma2020learning, hong2021garment4d}. Therefore, we must address the extreme deformation that manipulated garments could undergo. 

\begin{figure*}[th!]
  \centering
   \includegraphics[width=0.9\linewidth]{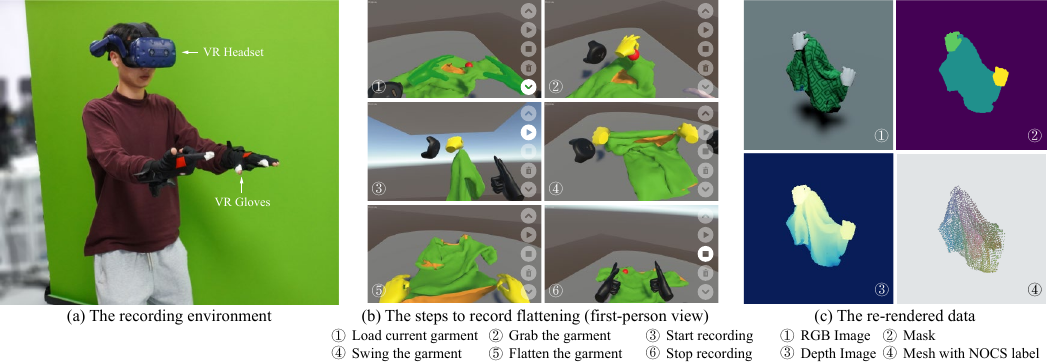}

   \caption{\textbf{The pipeline of our Virtual Realty recording system (VR-Garment).} (a) A volunteer needs to put on a VR headset and VR gloves. (b) By following the guidance of a specially designed UI, the volunteer begins to collect data efficiently.  (c) After recording, we re-render multi-view RGB-D images with Unity\cite{unity} and obtain masks and deformed garment meshes with NOCS labels.}
   \label{fig:vr_recording}
\end{figure*}

To tackle the garment pose tracking problem, we need a dataset of garment manipulation with complete pose annotations. However, such a dataset does not exist so far to the best of our knowledge. To build such a dataset, we turn to a VR-based solution due to the tremendous difficulty of garment pose annotation in the real world \cite{bozic2020deepdeform}. We first create a real-time VR-based recording system named \textbf{VR-Garment}. Then the volunteer can manipulate the garment in a simulator through the VR interface. 
With VR-Garment, we build a large-scale garment manipulation dataset called \textbf{VR-Folding}. Compared to the single static garment configuration (\ie grasped by one point) in GarmentNets, our manipulation tasks include \textit{flattening} and \textit{folding}, which contain much more complex garment configurations. In total, our VR-Folding dataset contains 9767 manipulation videos which consist of 790K multi-view RGB-D frames with full garment pose and hand pose annotations on four garment categories selected from the CLOTH3D \cite{bertiche2020cloth3d} dataset.

With the VR-Folding dataset, we propose an end-to-end online tracking method called \textbf{GarmentTracking} to perform category-level garment pose tracking during manipulation. For the garment pose modeling, we follow GarmentNets \cite{chi2021garmentnets} to adopt the normalized object coordinate space (NOCS) for each category.
Nevertheless, tracking garment pose raises new challenges compared to single-frame pose estimation: (1) How to fuse inter-frame geometry and correspondence information? (2) How to make the tracking prediction robust to pose estimation errors? (3) How to achieve tracking in real-time? To address these challenges, we conduct GarmentTracking in three stages, namely \textit{NOCS predictor}, \textit{NOCS refiner}, and \textit{warp field mapper}. Firstly, it predicts per-frame features and fuses them for canonical coordinate prediction. Then it refines the predicted canonical coordinates and the geometry with a NOCS refiner to reduce the accumulated errors. Finally, it maps the prediction in canonical space to the task space (\ie coordinate frame of the input point cloud).

Since no previous work is designed for the tracking setting, we use GarmentNets \cite{chi2021garmentnets} as a single-frame prediction baseline for comparison. 
We also perform extensive ablative experiments to reveal the efficacy of our design choices. Finally, we collect real-world data on garment manipulation and show the qualitative results of our method. In our design, we avoid the computationally expensive Marching Cubes \cite{marching_cubes} for reconstructing the canonical mesh frame by frame, so that we can achieve tracking at 15 FPS with an RTX 3090 GPU (\textbf{5 times faster} than the baseline approach).

We summarize our contributions as follows:

1). We propose a VR-based garment manipulation recording system named \textbf{VR-Garment}. It can synchronize human operations into the simulator and collect garment manipulation data.

2). We propose a large-scale garment manipulation dataset named \textbf{VR-Folding} for pose tracking. During manipulation, garments exhibit diverse configurations.

3). We propose a real-time end-to-end framework named \textbf{GarmentTracking} for category-level garment pose tracking. It can serve as a strong baseline for further research. We also demonstrate its generalization ability to real-world garment recordings with models trained by simulated data.

\section{Related Work}
\label{sec:related}
\textbf{Category-level Object Pose Estimation and Tracking.}
Object pose is the configuration of the object posited in the observation space. For the rigid object, we can describe its pose in 6 degrees of freedom (DOFs), \ie 3 for translation and 3 for rotation. However, for the non-rigid object, like garments, the object pose can be of near-infinite DOFs. 

On the other hand, the category-level object pose estimation task aims to learn a model that can predict unseen object poses of the same category \cite{wang2019normalized,li2020category,add1,add2,add3}. The concept is first introduced to estimate rigid object pose \cite{wang2019normalized}.
In \cite{wang2019normalized}, Wang \etal proposed a Normalized Object Coordinate Space (NOCS) as a category-specific canonical representation. 
Following the idea, Li \etal \cite{li2020category} proposed a hierarchical NOCS representation for articulated objects.

To handle the near-infinite DOF nature and the category-level generalization requirement, GarmentNets \cite{chi2021garmentnets} also defined NOCS for each garment category. They predicted the garment pose by mapping the reconstructed mesh from canonical space to task space.
However, GarmentNets \cite{chi2021garmentnets} treats each frame individually, which hampers its stability for inter-frame prediction, and its ability to infer complex poses from sequential movements. Our GarmentTracking is proposed for these tracking issues.

\textbf{Non-rigid Tracking and Reconstruction.}
Tracking and reconstructing non-rigid deforming objects is an important research area in computer vision and graphics. One line of works \cite{newcombe2015dynamicfusion, innmann2016volumedeform, guo2017real, slavcheva2017killingfusion, slavcheva2018sobolevfusion, bozic2020deepdeform, bozic2020neural} perform free-form tracking, which does not assume any geometric prior. For example, DynamicFusion \cite{newcombe2015dynamicfusion} used a hierarchical node graph structure and an efficient GPU solver to reconstruct the visible surface of the object. Deepdeform \cite{bozic2020deepdeform} and Bozic \etal \cite{bozic2020neural} leveraged learning-based correspondences to track deformable objects. However, unlike pose estimation methods which reconstruct the complete configuration of objects, all these methods can not reconstruct occluded parts.

Another line of works \cite{dou2016fusion4d, dou2017motion2fusion, li20214dcomplete, niemeyer2019occupancy, palafox2021npms, palafox2022spams, lin2022occlusionfusion} can perform 4D reconstruction from RGB-D videos, which captures the complete geometry of the object both in space and time. Unfortunately, the shape representations of these methods have limitations when adapting to garment pose reconstruction under large deformations. For example, Fusion4D \cite{dou2016fusion4d}, Motion2fusion \cite{dou2017motion2fusion}, 4DComplete \cite{li20214dcomplete}, OcclusionFusion \cite{lin2022occlusionfusion}, NPMs \cite{palafox2021npms} and OccupancyFlow \cite{niemeyer2019occupancy} heavily rely on watertight object modeling such as SDF, TSDF, or occupancy grids to reconstruct object surfaces. Such modeling is not suitable for objects with open and thin structures like garments.

\textbf{Garment-related Dataset.}
Current garment-related datasets can be divided into \textit{asset datasets} \cite{bertiche2020cloth3d, zhu2020deep} and \textit{task datasets} \cite{bozic2020deepdeform, verleysen2020video, ma2020learning, chi2021garmentnets}. \textit{Asset datasets} provide garment models for different tasks. For example, GarmentNets \cite{chi2021garmentnets} proposed a simulation dataset for category-level pose-estimation task based on CLOTH3D \cite{bertiche2020cloth3d}. We also build our VR-Folding dataset based on \cite{bertiche2020cloth3d}.
Other task datasets do not require complete garment models \cite{ma2020learning}. For example, CAPE \cite{ma2020learning} deals with the clothed human reconstruction task. However, the human body limits the possible garment states.
DeepDeform \cite{bozic2020deepdeform} dataset contains simple scenes where a person lifts one garment with minor deformations, and it only annotates sparse keypoint correspondences between frames. 
A real-world cloth-folding dataset proposed by Verleysen \etal \cite{verleysen2020video} contains videos of cloth-folding actions, but it only annotates the contour of garments in 2D images.
Our VR-Folding dataset is the first dataset designed for category-level garment pose tracking in manipulation, and it contains dynamic scenes which include complex human actions and garment configurations. 

\begin{figure*}[h!]
  \centering
   \includegraphics[width=\linewidth]{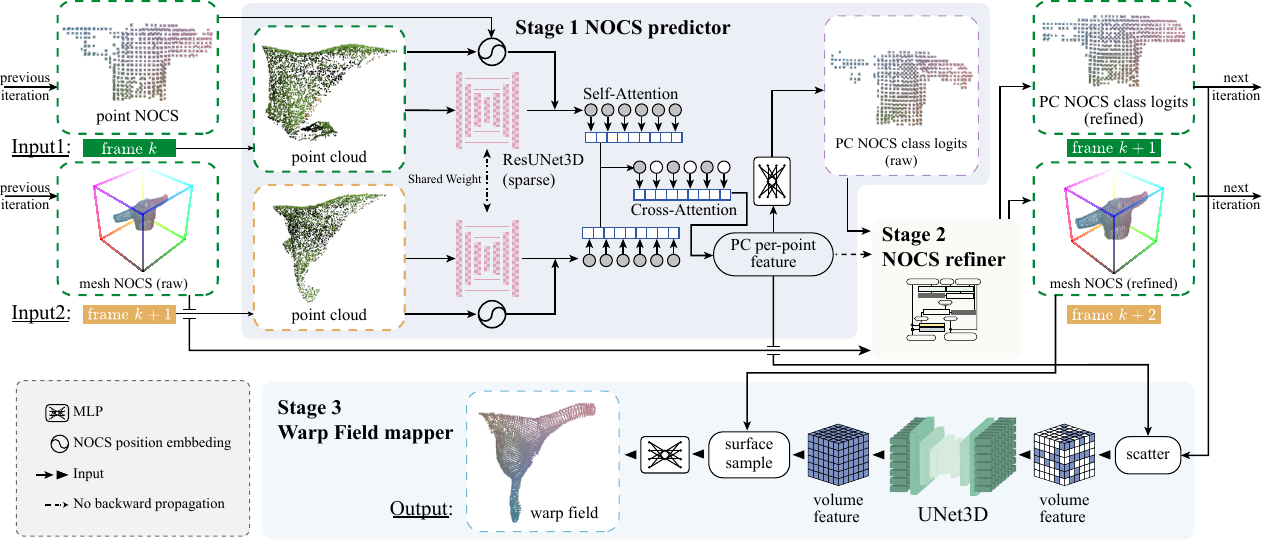}

   \caption{The overview of \textbf{GarmentTracking}. Given the per-point NOCS coordinate of the first frame and a rough canonical shape (mesh NOCS), our tracking method takes two frames of the partial point cloud as input. In stage 1, the NOCS predictor will generate an inter-frame fusion feature and predict raw NOCS coordinates. In stage 2, the NOCS refiner will refine the NOCS coordinates and the canonical shape simultaneously. In stage 3, the warp field mapper will predict the warp field which maps from canonical space to task space.}
   \label{fig:network_overview}
\end{figure*}

\section{VR-Folding Dataset} 
\label{sec:dataset}

To build the VR-Folding dataset, we develop a real-time data recording system called \textbf{VR-Garment} (\cref{fig:vr_recording}). In this way, we can combine human experience and the benefits of simulation environments (\ie easy access to ground-truth poses) to efficiently collect a large amount of data with natural and complex poses. We will describe the system design and the operation procedures in Sec. \ref{sec:vr_garment} and Sec. \ref{sec:task_definition}. Then we will describe the data statistics in Sec. \ref{sec:dataset_stat}. 

\subsection{VR-Garment}\label{sec:vr_garment}
In this section, we will first describe the hardware and software setup of the system and then the recording procedure for volunteers to operate. The recording system is illustrated in Fig. \ref{fig:vr_recording}.
    
    \textbf{Hardware.} On the hardware side, our recording system needs an HTC Vive Pro \cite{htcVicePro} VR headset and Noitom Hi5 \cite{NoitomHi5} VR gloves which can track finger poses in the real world and reproduce them with virtual hands in the simulator through a VR interface.
    
    \textbf{Software.} 
    We developed our VR recording framework based on Unity \cite{unity} for its good support of mainstream VR devices. 
    The cloth physics simulation in Unity is achieved by Obi \cite{obi}. 
    Specifically, we implemented a simple UI and a grasping system in VR, which allow users to grasp or release any point on the garment surface when two virtual fingertips make contact with the garment.
    
    \textbf{Recording Procedure.} 
    Firstly, a volunteer must wear a VR headset and VR gloves. Then he should observe a garment instance randomly dropping on a table in Unity. Next, he performs a pre-defined manipulation task (\eg folding). In the manipulation process, we save the deformed garment mesh and hand poses for each frame. When the task is done, the volunteer will use special gestures (\eg fist) to send commands for moving to the next garment instance. After recording, we re-render multi-view RGB-D images in Unity with the saved animation data and generate corresponding ground-truth annotations (\eg garment poses, masks). Note that all the rendering settings (\eg lights, cameras, textures \etc) can be customized even after recording. 
    
\subsection{Task Definition}\label{sec:task_definition}
    In a typical cloth folding process, we operate this task in two stages, namely \textit{flattening} and \textit{folding}: 
    
    \textbf{Flattening:} Firstly, a garment will drop on the virtual table in Unity. Then our system will randomly choose one point on the garment surface. Next, the volunteer will grasp that point with one hand and lift the garment in the air (an initial configuration similar to that in GarmentNets \cite{chi2021garmentnets}). Next, the volunteer will try to grasp and fling the garment repeatedly with two hands until the garment is smoothed and in a flattened T-pose (see \cref{fig:vr_recording}). Please see the supplementary files for more details about the task.
    
    \textbf{Folding:} Firstly, a garment in a flattened T-pose will be placed on the virtual table. Then the volunteer will repeatedly perform pick-and-place actions with both hands until the garment is folded. Though people may have different preferred steps to achieve folding, we have defined general rules for each category and asked the volunteers to follow them. Please see the supplementary files for more details.

\subsection{Data Statistics}\label{sec:dataset_stat}
All the garment meshes used in our system are from CLOTH3D \cite{bertiche2020cloth3d}. We choose 4 categories from CLOTH3D dataset, namely \textit{Shirt}, \textit{Pants}, \textit{Top} and \textit{Skirt}. 
For \textit{flattening} task, we recorded 5871 videos which contain 585K frames in total. For \textit{folding} task, we recorded 3896 videos which contain 204K frames in total. As shown in \cref{fig:vr_recording}, the data for each frame include multi-view RGB-D images, object masks, full garment meshes, and hand poses. Please see the supplementary files for more statistics on the dataset.

\section{Method}
\label{sec:method}
This paper proposes an end-to-end online tracking method called \textbf{GarmentTracking} for category-level garment pose tracking. 
As shown in \cref{fig:network_overview}, given a first-frame garment pose (point NOCS, \ie canonical coordinates of partial point cloud) and a rough canonical shape (mesh NOCS, \ie sampled points from mesh in canonical space) of an instance, it takes point cloud sequences as input and generates complete garment geometry with inter-frame correspondence (\ie NOCS coordinates). Specifically, GarmentTracking can be divided into three stages. 
In the first stage (\cref{sec:canonical_pred}), the network will predict canonical coordinates for the partial input point cloud. In the second stage (\cref{sec:refiner}), the network will refine the predicted canonical coordinates and the input canonical shape. In the third stage (\cref{sec:warping}), the network will use the refined canonical shape and canonical coordinates to predict a warp field that maps from canonical space to task space (\ie the coordinate frame of the input point cloud).

\subsection{Canonical Coordinate Prediction}
\label{sec:canonical_pred}
\subsubsection{Normalized Garment Canonical Space}
Following the definition of garment representation in GarmentNets \cite{chi2021garmentnets}, we use Normalized Object Coordinate Space (NOCS) coordinates as an intermediate representation for object states in a category. As shown in \cref{fig:network_overview}, the rest state of a garment is the T-pose defined by the garment worn by a person (provided by CLOTH3D \cite{bertiche2020cloth3d} dataset).

\subsubsection{3D Feature Extractor}
The thin structure and near-infinite DOF nature of garments may result in many complicated poses (\eg multi-layer cloth stacked together) that require feature extraction for granular local details. 
Our method uses a high-resolution sparse 3D convolution network (ResUNet3D) proposed by FCGF \cite{choy2019fully} to extract the per-point feature from the raw point cloud. ResUNet3D is a UNet-like network with skip connections and residual blocks. 
Please refer to the supplementary files for further details of the network.

\subsubsection{Inter-frame Feature Fusion with Transformer}
After extracting the feature from the extractor, we apply the inter-frame feature fusion with Transformer \cite{vaswani2017attention}.

\textbf{Feature Matching}
Inspired by PTTR \cite{zhou2022pttr}, we perform feature matching with self-attention and cross-attention modules based on Transformer \cite{vaswani2017attention}. In general, we first use a self-attention module to individually aggregate point features for the two input frames. Then we use a cross-attention module to perform feature matching between two frames. Intuitively, the self-attention operation can have a global understanding of the current frame, and the cross-attention operation can capture cross-frame correlations and generate a relation-enhanced fusion feature. The self-attention and cross-attention modules are based on the relation attention module (RAM) proposed by PTTR \cite{zhou2022pttr}. 

Please see the supplementary files for more details on the relation attention module. 

\textbf{NOCS Prediction} After obtaining the per-point fusion feature via the cross-attention module, we predict the per-point canonical coordinate with MLP. We follow GarmentNets \cite{chi2021garmentnets} and formulate this problem as a classification task instead of a regression task. Specifically, we divide each axis into 64 bins and the network independently predicts each axis's classification score.
During training, we use a cross-entropy loss to supervise the classification scores.

\textbf{NOCS Coordinates for Positional Embedding}\label{sec:pos_emb}
We have per-point NOCS coordinate prediction from the previous frame, which contains clearer geometric and structural information. We use it for positional embedding \cite{vaswani2017attention}, which will be added to input features before feeding into transformers. The positional embeddings for two frames are calculated as \cref{eq:pos_emb}: 
\begin{equation}\label{eq:pos_emb}
    \mathbf{emb}_1 = f_1([\mathbf{P}_1^{xyz}, \mathbf{P}_1^{nocs}]), \mathbf{emb}_2 = f_2([\mathbf{P}_2^{xyz}]),
\end{equation}
where $\mathbf{P}_1^{xyz}$ and $\mathbf{P}_2^{xyz}$ are the partial input point clouds of the two frames, and $\mathbf{P}_1^{nocs}$ is the predicted per-point NOCS coordinates of the partial point cloud in the previous frame. Here $f_1(\cdot)$ and $f_2(\cdot)$ are learned MLP. By fusing NOCS coordinates into positional embedding, the transformer network will incorporate positional and semantic information from previous frames. Besides, empirically speaking, utilizing intermediate representations like NOCS coordinates instead of complete garment poses can increase the robustness against noisy predictions during long-term tracking.
\subsection{NOCS Refiner}
\begin{figure}[h!]
  \centering
   \includegraphics[width=0.8\linewidth]{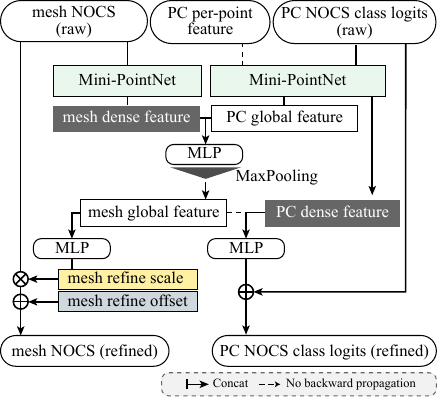}
   \caption{PC-Mesh Fusion Refiner} 
   \label{fig:refiner}
\end{figure}
\label{sec:refiner}
Since the canonical shape can be generated by other methods like GarmentNets \cite{chi2021garmentnets}, or augmented with noise, it might be inaccurate. On the other hand, the NOCS coordinate predictions can also be noisy. Such inaccuracy could cause errors to be accumulated during tracking. To mitigate this problem, we propose a NOCS PC (Point Cloud)-Mesh intertwined refiner, \textbf{NOCS refiner}. As shown in \cref{fig:refiner}, the predicted NOCS coordinates can reveal the cues of the input point cloud, such as scales and offsets, while the canonical shape can provide information about the complete geometry. Thus they can complement each other. We describe the NOCS refiner in two parts (\textit{PC refiner} and \textit{Mesh refiner}):
 
 \textbf{PC Refiner:} Firstly, the predicted NOCS classification scores and the per-point fusion feature from the transformer will be concatenated and fed into a Mini-Pointnet \cite{qi2017pointnet}. Next, the dense feature will be fused with the global mesh feature generated by \textit{Mesh Refiner} with concatenation. Finally, we use MLP to predict the final delta logits with the fused dense feature. We use cross-entropy loss to supervise the refined classification logits during training.
 
 \textbf{Mesh Refiner:}  Firstly, we use a Mini-Pointnet to extract dense features from raw canonical shapes. Then we concatenate the dense mesh feature generated by \textit{PC Refiner} with the global feature from the partial point cloud to obtain the fused dense feature. Next, we use an MLP with global pooling to extract the final global shape feature. Finally, we predict the global scale factor and offset for the canonical shape with the global shape feature and an MLP. Finally, we use L2 loss to supervise the refined mesh points during training.

\subsection{Warping from Canonical To Task Space}
\label{sec:warping}
\subsubsection{Feature Scattering with Canonical Coordinates}
After obtaining the refined canonical (NOCS) coordinate prediction (\cref{sec:refiner}) of a partial point cloud, we scatter the per-point feature generated by Transformer (\cref{sec:canonical_pred}) into a $32^3$ feature volume. The ``scatter'' operation is performed by copying the feature vector to the target location in volume with predicted NOCS coordinates. All features mapped to the same volume index will be aggregated with a channel-wise maximum operation. And all the volume locations with no corresponding feature vectors are filled with zeros. Then the feature volume will be fed into a 3D UNet \cite{cciccek20163d}, then we can obtain a dense feature volume $\mathcal{V}$ for warp field prediction.

\subsubsection{Warp Field Prediction}\label{sec:warp_field}
Finally, we map the refined canonical shape (\cref{sec:refiner}) from canonical space to task space. The output contains the full configuration of the garment, including the occluded parts. It is achieved by warp field prediction\cite{chi2021garmentnets}, which is an implicit neural function $w(p;\mathcal{V})\in \mathbf{R}^3$ that takes a query point $p$ in the canonical space as input and infers the corresponding location of $p$ in task space. Here $w(\cdot)$ is a learned MLP. We use L2 loss to supervise the warp field prediction. In training, the query points are sampled from the canonical mesh surface. In inference, the query points are generated by our Mesh Refiner (\cref{sec:refiner}).

\section{Experiments}
\label{sec:exp}
\subsection{Implementation Details}\label{sec:impl_detail}
We implement our method with Pytorch \cite{pytorch} and use Adam optimizer with a learning rate of 0.0001. The training stage takes about 150 epochs to converge, which lasts for 1-3 days on an RTX 3090 GPU, depending on the training dataset sizes for different categories. We randomly sample 4000 points from the input partial point cloud and 6000 points from the input canonical mesh surface for each frame. During training, we randomly add noise to the partial point-cloud canonical coordinates $\mathbf{P}_1^{nocs}$ of the previous frame by randomly generating a NOCS scale factor $\mathbf{s}_{pc}\in[0.8,1.2]^3$ and a global NOCS offset $\mathbf{o}_{pc}\in[0, 0.1]^3$. We also add noise to the input canonical mesh by randomly generating a global NOCS scale factor $\mathbf{s}_{mesh}\in[0.8,1.2]^3$ during training. Please see the supplementary files for further details on training, inference, and network structure. 
\subsection{Metrics}

\textbf{NOCS Coordinate Distance}~($D_{nocs}$). We calculate the point-wise L2 distance between the predicted NOCS coordinate of the partial point cloud with the ground-truth NOCS labels. This metric evaluates the quality of per-point NOCS coordinate prediction for input partial point cloud. 

\textbf{Chamfer Distance}~($D_{chamf}$). We calculate the Chamfer distance in centimeters between the reconstructed mesh points and the ground-truth mesh points in task space.
    This metric can evaluate the quality of \textit{surface reconstruction}.

\textbf{Correspondence Distance}~($D_{corr}$, $A_d$). We calculate point-wise L2 distance in centimeters between the reconstructed mesh and the ground-truth mesh for each frame in task space. The correspondences are based on the NOCS coordinates (\ie each point on the predicted mesh will find the closest point on the ground-truth mesh in NOCS). This metric can evaluate the quality of garment \textit{pose estimation}. In practice, we find the variance of the error distribution in different frames is very large, which makes the mean correspondence distance $D_{corr}$ across all frames dominated by the worst cases. So we additionally introduce $A_d$ which represents the accuracy (\ie ratio of frames) with $D_{corr}<d$.

\subsection{Experiment Results}
\subsubsection{Main Experiments}\label{sec:main_exp}
\noindent\textbf{Baselines}\label{sec:baselines}
In \cref{tab:main_exp}, we compare GarmentNets with two settings of our method:

\textbf{GarmentNets} \cite{chi2021garmentnets}: As the only-existing method for category-level garment pose estimation, GarmentNets focused on the single-frame setting. We adapt it for tracking by prediction frame by frame. 

\textbf{Ours (GT)}: Our tracking method given the ground-truth first-frame garment pose and the ground-truth canonical mesh as initialization.

\textbf{Ours (Pert.)}: Our tracking method when the first-frame garment pose and the input canonical shape are perturbed with noise. Specifically, we use the same noise distribution in training (\cref{sec:impl_detail}) which adds global NOCS scale and offset to the first-frame canonical coordinates of partial point-cloud and the canonical mesh. Additionally, we add per-point Gaussian noise ($\delta$=0.05) to the input canonical coordinates of the first frame during inference.

\begin{table*}[h!]
\resizebox{\textwidth}{!}{
  \centering
  \begin{tabular}{c|c|c|ccc|c|c|ccc|c|c}
    \toprule
    \multicolumn{1}{c|}{\multirow{2}{*}{Type}} & \multicolumn{1}{c|}{\multirow{2}{*}{Method}} & \multicolumn{1}{c|}{\multirow{2}{*}{Init.}} & \multicolumn{5}{c|}{Folding} & \multicolumn{5}{c}{Flattening}\\ \cline{4-13}
    & & & $A_{3cm}\uparrow$ & $A_{5cm}\uparrow$ & $D_{corr}\downarrow$ & $D_{chamf}\downarrow$ & $D_{nocs}\downarrow$ & $A_{5cm}\uparrow$ & $A_{10cm}\uparrow$ & $D_{corr}\downarrow$ & $D_{chamf}\downarrow$ & $D_{nocs}\downarrow$ \\
    \midrule
    \multirow{3}{*}{Shirt} & GarmentNets \cite{chi2021garmentnets} & N/A & 0.8\% & 21.5\% & 6.40 & 1.58 & 0.221 & 13.2\% & 59.4\% & 10.54 & 3.54 & 0.135\\ 
    & Ours & GT & \textbf{29.8\%} & 85.8\% & \textbf{3.88} & \textbf{1.16} & \textbf{0.051} & \textbf{30.7\%} & \textbf{83.4\%} & \textbf{8.63} & \textbf{1.78} & \textbf{0.105} \\ 
    & Ours & Pert. & 29.0\% & \textbf{85.9\%} & 3.88 & 1.18 & 0.052 & 25.4\% & 81.6\% & 8.94 & 1.85 & 0.109 \\ \hline
    \multirow{3}{*}{Pants} & GarmentNets \cite{chi2021garmentnets} & N/A & 16.2\% & 69.5\% & 4.43 & 1.30 & 0.162 & 1.5\% & 42.4\% & 12.54 & 4.19 & 0.185\\ 
    & Ours & GT & \textbf{47.3\%} & \textbf{94.0\%} & \textbf{3.26} & \textbf{1.07} & \textbf{0.039} & \textbf{31.3\%} & \textbf{78.2\%} & \textbf{8.97} & \textbf{1.64} & \textbf{0.113}\\ 
    & Ours & Pert. & 42.8\% & 93.6\% & 3.35 & 1.10 & 0.039 & 30.7\% & 76.9\% & 9.55 & 2.71 & 0.143 \\ \hline
    \multirow{3}{*}{Top} & GarmentNets \cite{chi2021garmentnets} & N/A & 10.3\% & 53.8\% & 5.19 & 1.51 & 0.148 &21.6\% & 57.6\% & 9.98 & 2.13 & 0.174\\ 
    & Ours & GT & \textbf{37.9\%} & \textbf{85.9\%} & \textbf{3.75} & \textbf{0.99} & \textbf{0.051} & \textbf{36.5\%} & \textbf{69.0\%} & \textbf{9.41} & \textbf{1.59} & \textbf{0.113} \\ 
    & Ours & Pert. &36.6\% & 86.1\% & 3.76 & 1.00 & 0.051 & 33.5\% & 68.1\% & 9.61 & 1.62 & 0.116 \\ \hline
    \multirow{3}{*}{Skirt} & GarmentNets \cite{chi2021garmentnets} & N/A & 1.1\% & 30.3\% & 6.95 & 1.89 & 0.239 & 0.1\% & 7.9\% & 18.48 & 5.99 & 0.287\\ 
    & Ours & GT & \textbf{23.5\%} & \textbf{71.3\%} & \textbf{4.61} & \textbf{1.33} & \textbf{0.060} & \textbf{5.4\%} & \textbf{39.4\%} & \textbf{16.09} & \textbf{2.02} & \textbf{0.199} \\ 
    & Ours & Pert. & 22.8\% & 70.6\% & 4.72 & 1.36 & 0.060 & 2.3\% & 35.5\% & 16.55 & 2.15 & 0.207\\ 
    \bottomrule
  \end{tabular}
}
  \caption{Quantitative results on VR-Folding dataset.}
  \label{tab:main_exp}
\end{table*}
\noindent\textbf{Results} 
\cref{tab:main_exp} summarizes the quantitative results on the VR-Folding dataset. 
In general, our method outperforms GarmentNets in all metrics by a large margin. 
On the challenging $A_{3cm}$ metric in \textit{Folding} task and $A_{5cm}$ in \textit{Flattening} task, GarmentNets has very low performance (\eg $0.8\%$ in \textit{Shirt Folding}), while our method achieves much higher scores (\eg $29.0\%$ in \textit{Shirt  Folding}), which proves that our method can generate more accurate predictions in videos compared to GarmentNets. Our method also outperforms GarmentNets on mean correspondence distance $D_{corr}$ and chamfer distance $D_{chamf}$, which proves that our method can do well in both \textit{pose estimation} and \textit{surface reconstruction} tasks. Even with perturbation on first-frame poses (Ours with Pert. in \cref{tab:main_exp}), our method only shows minor performance loss (\eg $37.9\%\rightarrow36.6\%$ in \textit{Top Folding}) compared to using ground-truth as first-frame pose.

\begin{figure}[h!]
  \centering
   \includegraphics[width=1\linewidth]{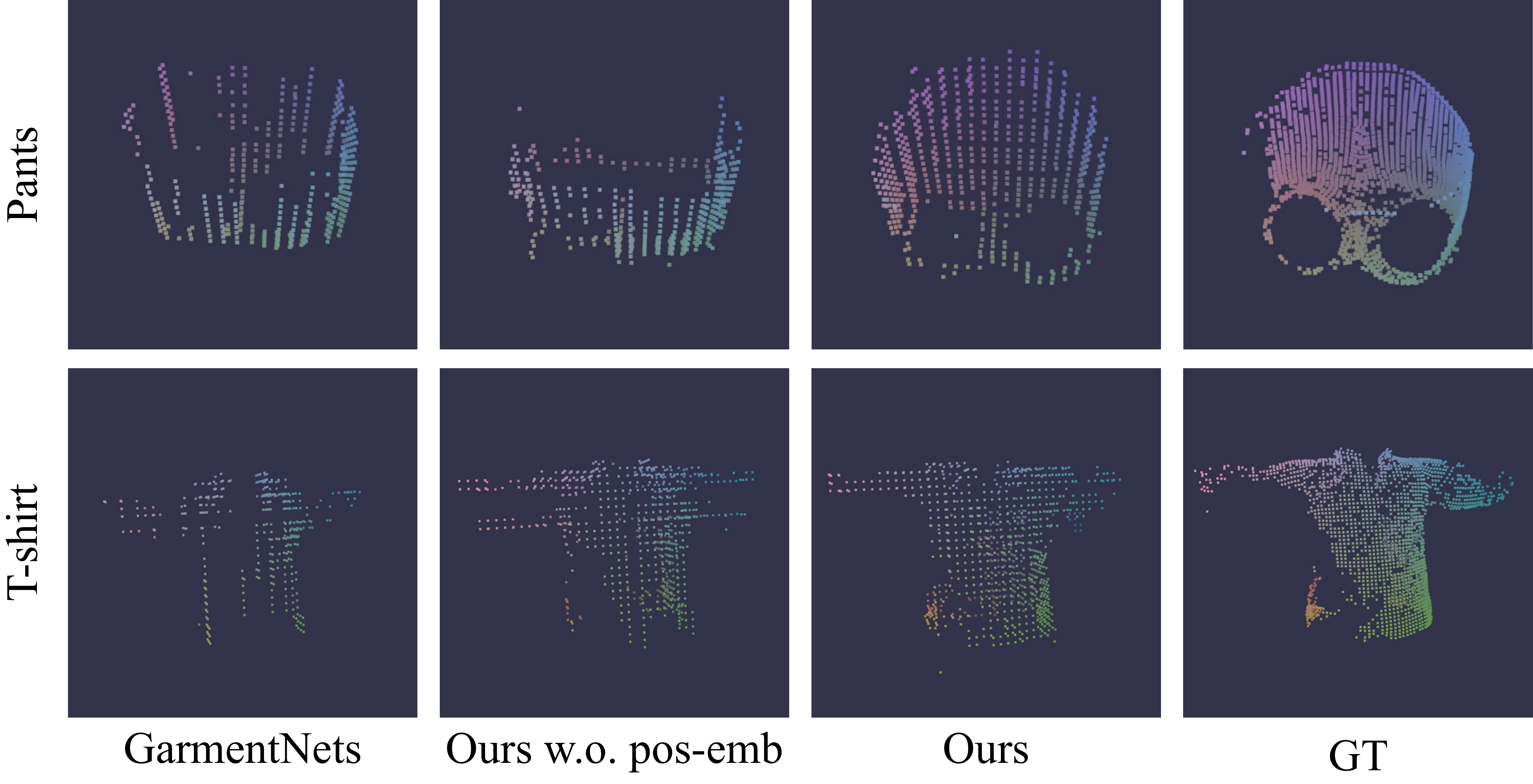}

   \caption{The canonical coordinate prediction results on the VR-Folding dataset.}
   \label{fig:pc_nocs}
   \vspace{-0.5cm}
\end{figure}

\begin{figure*}[h!]
  \centering
   \includegraphics[width=0.8\linewidth]{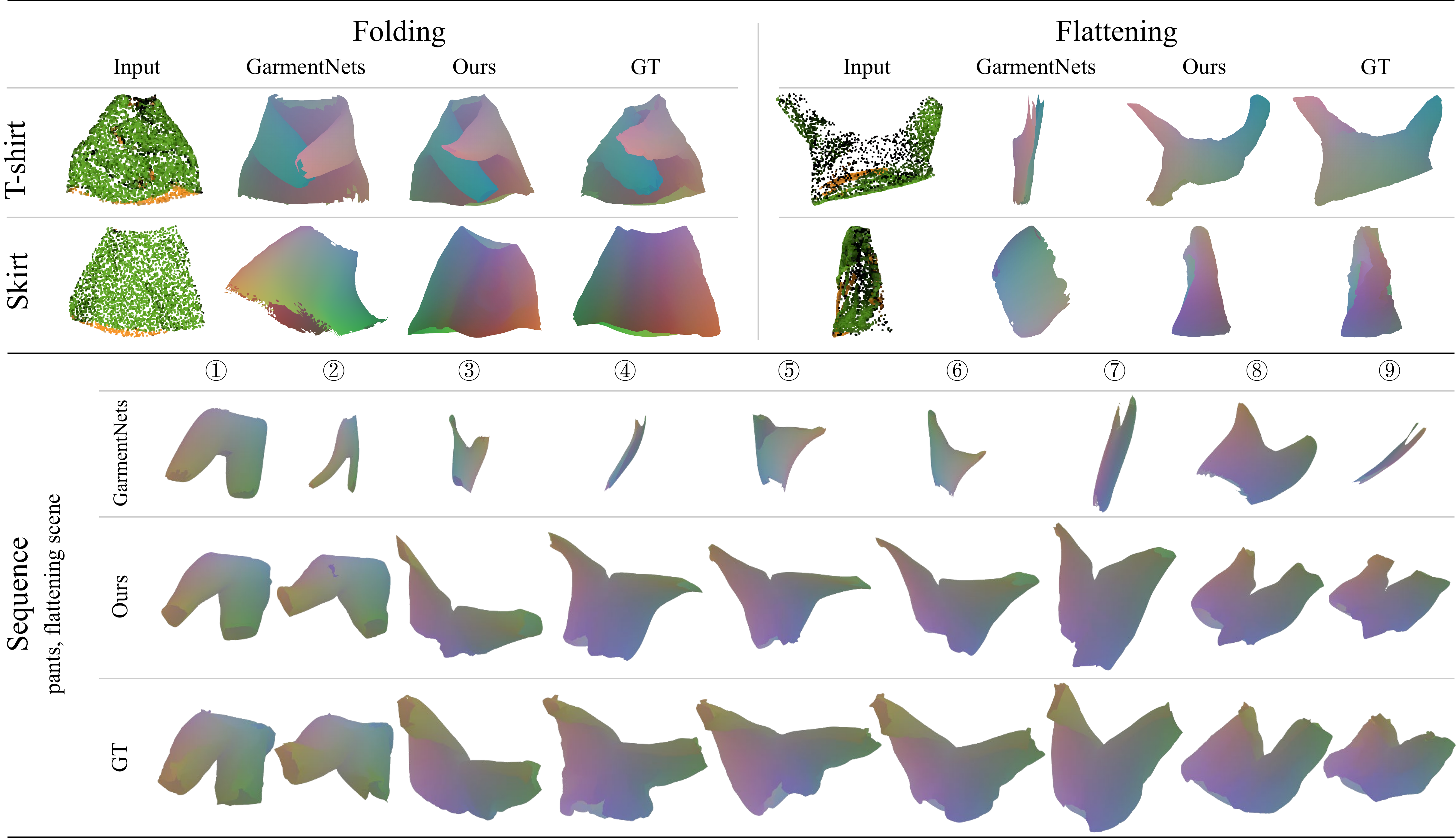}

   \caption{The qualitative results of pose estimation for \textbf{unseen} instances in VR-Folding dataset. 
   In the long sequence tracking (shown in the lower part), our prediction still keeps high consistency with GT, while GarmentNets outputs a series of meshes that lack stability.}
   \label{fig:qualitative}
   \vspace{-0.5cm}
\end{figure*}

We also present some qualitative results in \cref{fig:pc_nocs} and \cref{fig:qualitative}. We can see from \cref{fig:qualitative} that the prediction results of GarmentNets are very unstable because it performs mesh reconstruction for each frame individually and can not utilize the information from previous frames. Conversely, our method can leverage input canonical mesh and inter-frame information to predict more stable and accurate results. Besides, GarmentNets suffers from ambiguity brought by symmetry (\eg take a front side as a back side), which hampers its ability to predict accurate canonical coordinates (see \cref{fig:pc_nocs}). In contrast, our method can predict much more accurate canonical coordinates (\eg $D_{nocs}$ 0.162 v.s. 0.039 for \textit{Pants Folding} in \cref{tab:main_exp}).

\subsubsection{Ablation Study}\label{sec:ablative}
\textbf{NOCS Positional Embedding.} In our method, NOCS positional embedding (\cref{sec:pos_emb}) is the crucial design choice to leverage inter-frame correspondence information. As shown in \cref{fig:pc_nocs}, if we remove the NOCS positional embedding from our network, the network will suffer from the same ambiguity problem as GarmentNets due to symmetry.

\begin{table*}[h!]
\resizebox{\textwidth}{!}{
  \centering
  \begin{tabular}{c|c|ccc|c|c|ccc|c|c}
    \toprule
    \multicolumn{1}{c|}{\multirow{2}{*}{Type}} & \multicolumn{1}{c|}{\multirow{2}{*}{Method}} & \multicolumn{5}{c|}{Folding} & \multicolumn{5}{c}{Flattening}\\ \cline{3-12}
    & & $A_{3cm}\uparrow$ & $A_{5cm}\uparrow$ & $D_{corr}\downarrow$ & $D_{chamf}\downarrow$ & $D_{nocs}\downarrow$ & $A_{5cm}\uparrow$ & $A_{10cm}\uparrow$ & $D_{corr}\downarrow$ & $D_{chamf}\downarrow$ & $D_{nocs}\downarrow$ \\
    \midrule
    \multirow{5}{*}{Shirt} & Ours & \textbf{29.0\%} & \textbf{85.9}\% & \textbf{3.88} & \textbf{1.18} & 0.052 & \textbf{25.4\%} & \textbf{81.6\%} & \textbf{8.94} & \textbf{1.85} & \textbf{0.109} \\ 
    & Ours w.o. PC refiner & 4.1\% & 25.3\% & 7.76 & 1.70 & 0.115 & 0.6\% & 27.4\% & 17.42 & 2.89 & 0.229\\ 
    & Ours w.o. Mesh refiner & 26.8\% & 83.6\% & 3.92 & 1.19 & \textbf{0.048} & 23.5\% & 81.2\% & 9.18 & 1.88 & 0.110\\
    & Ours w. PointNet++ \cite{qi2017pointnet++} & 2.2\% & 34.7\% & 6.53 & 1.54 & 0.085 & 13.3\% & 53.2\% & 14.21 & 1.93 & 0.174\\
    \hline
    \multirow{5}{*}{Pants} & Ours & \textbf{42.8}\% & \textbf{93.6\%} & \textbf{3.35} & \textbf{1.10} & \textbf{0.039} & \textbf{30.7\%} & \textbf{76.9\%} & \textbf{9.55} & \textbf{2.71} & 0.143 \\ 
    & Ours w.o. PC refiner & 23.1\% & 70.0\% & 4.84 & 1.30 & 0.072 & 5.8\% & 49.2\% & 13.72 & 3.46 & 0.165\\
    & Ours w.o. Mesh refiner & 33.5\% & 92.2\% & 3.52 & 1.18 & 0.039 & 22.5\% & 75.2\% & 9.76 & 2.78 & 0.148\\
    & Ours w. PointNet++ \cite{qi2017pointnet++} & 8.9\% & 73.6\% & 4.91 & 1.33 & 0.066 & 8.0\% & 69.1\% & 10.12 & 2.81 & \textbf{0.125}\\
    \bottomrule
  \end{tabular}
}
  \caption{Results of ablative experiments on VR-Folding dataset.}

  \label{tab:ablation}
  \vspace{-0.5cm}
\end{table*}

\textbf{NOCS Refiner.} 
Unlike rigid object tracking, garment tracking has a higher demand for avoiding error accumulation in long videos, because the error distribution of predicted pose in testing can be very different from that in training due to the near-infinite DOF.
As shown in \cref{tab:ablation}, our proposed PC Refiner (\cref{sec:refiner}) greatly influences the performance due to its ability to refine NOCS coordinate predictions in each frame. Besides, the Mesh Refiner (\cref{sec:refiner}) also contributes to a slight performance improvement, indicating that our network has more tolerance for canonical mesh errors than NOCS coordinate errors.

\textbf{Feature Extractor.} 
As shown in \cref{tab:ablation}, after we replace our feature extractor (\ie ResUNet3D \cite{choy2019fully} based on sparse 3D convolution) with PointNet++ \cite{qi2017pointnet++}, the overall performance drops a lot.
Thus high-resolution 3D convolution network should be a better choice for this task.

\subsubsection{Robustness}
\begin{figure}[h!]
  \centering
   \includegraphics[width=1\linewidth]{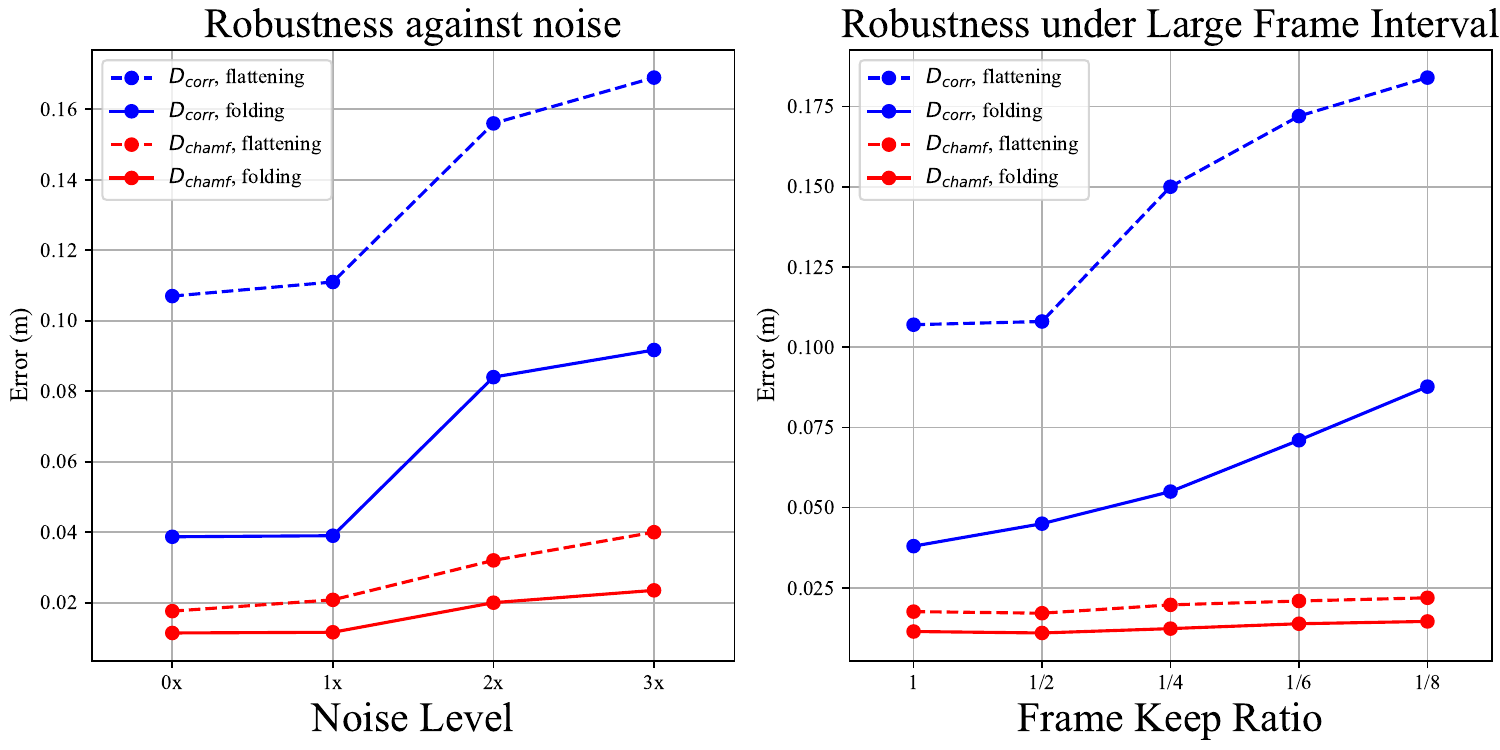}

   \caption{The robustness experiments.}
   \label{fig:robustness}
   \vspace{-0.7cm}
\end{figure}

\textbf{Robustness against Noise}
We test our method under different levels of noise perturbation by increasing the initial pose noise level described in \cref{sec:main_exp} by 1 or 2 times. Specifically, we augment the point-cloud NOCS coordinates of the first frame with a global scaling factor $s_{pc}$, a global offset $o_{pc}$, and Gaussian noise standard deviation $\delta$. Besides, we also augment the canonical mesh with global scaling factor $s_{mesh}$.  As shown in \cref{fig:robustness} (left), We can see that our method performs well at \textit{surface reconstruction} (\ie $D_{chamf}$) under high noise level. Please see the supplementary files for details of the noise parameters.

\textbf{Robustness under Large Frame Interval}
Tracking at large frame intervals with huge object deformations can be very challenging. Therefore, we uniformly drop frames in videos and only keep part (\ie 1/2, 1/4, 1/6, and 1/8) of the frames. The results are shown in  \cref{fig:robustness} (right), we can see that our method is more robust against missing frames on \textit{Folding} task compared to \textit{Flattening} task.

\subsubsection{Tracking Speed}
On a single RTX 3090 GPU, GarmentNets takes 100ms to pass the backbone, 7ms for volume query, and 170ms for Marching Cubes \cite{marching_cubes}. It results in a runtime of 3.6 FPS. While in our design, we adopt a faster backbone that costs only 45ms and eliminates the time-consuming Marching Cubes. Our NOCS refiner and warp field prediction take 12ms and 7ms respectively. Our method can achieve 15 FPS during inference, which is $\sim 5$ times faster than \cite{chi2021garmentnets}.

\subsubsection{Generalization Ability}
\textbf{Neural Prediction as First-Frame Pose}
In order to evaluate the generalization ability of our method, we directly use GarmentNets prediction (\ie canonical coordinates and mesh) as the first-frame pose during inference. As shown in \cref{tab:neural_pred}, our method still outperforms GarmentNets by a large margin without any data augmentation related to GarmentNets during training.

\begin{table}[h!]
\resizebox{0.48\textwidth}{!}{
  \centering
  \begin{tabular}{c|c|ccc|c|c}
    \toprule
    Type & Method &
    $A_{3cm}\uparrow$ & $A_{5cm}\uparrow$ & $D_{corr}\downarrow$ & $D_{chamf}\downarrow$ & $D_{nocs}\downarrow$\\
    \midrule
    \multirow{2}{*}{Shirt} & Ours* & \textbf{25.4\%} & \textbf{78.9\%} & \textbf{4.04} & \textbf{1.18} & \textbf{0.052}\\
    & GarmentNets & 0.8\% & 21.5\% & 6.40 & 1.58 & 0.221\\ \hline
    \multirow{2}{*}{Pants} & Ours* & \textbf{45.1\%} & \textbf{92.2\%} & \textbf{3.33} & \textbf{1.16} & \textbf{0.040}\\ 
    & GarmentNets & 16.2\% & 69.5\% & 4.43 & 1.30 & 0.162\\ \hline
    \multirow{2}{*}{Top} & Ours* & \textbf{21.1\%} & \textbf{61.9\%} & \textbf{4.82} & \textbf{1.11} & \textbf{0.065}\\ 
    & GarmentNets & 10.3\% & 53.8\% & 5.19 & 1.51 & 0.148 \\ \hline
    \multirow{2}{*}{Skirt} & Ours* & \textbf{14.7\%} & \textbf{65.9\%} & \textbf{5.36} & \textbf{1.46} & \textbf{0.078}\\ 
    & GarmentNets & 1.1\% & 30.3\% & 6.95 & 1.89 & 0.239 \\ 
    \bottomrule
  \end{tabular}
}
  \caption{Results of our method using GarmentNets prediction as first-frame pose on \textit{Folding} task.}
  \label{tab:neural_pred}
  \vspace{-0.2cm}
\end{table}

\textbf{Real World Experiments}
We collect some real-world RGB-D videos of garment manipulation with Realsense L515\cite{L515} LiDAR cameras. Our method can directly track garment pose for novel garments in the real-world with a model trained only on our simulated data. Please see the supplementary files for more qualitative results.
\section{Conclusion and Future Works}
In this work, we propose a complete framework for garment pose tracking, including the data collection (\ie VR-Garment system), dataset (\ie VR-Folding), and a strong approach (\ie GarmentTracking) which is both quantitatively and qualitatively better than the baseline approach. As a platform, we believe VR-Garment can innovate the dataset collection for other kinds of deformable objects. As a manipulation dataset, we are interested in using VR-Folding for robot imitation learning. As a strong baseline, we hope GarmentTracking can facilitate future research in this challenging direction.

\section*{Acknowledgement}
This work was supported by the National Key Research and Development Project of China
(2021ZD0110704), Shanghai Municipal Science and
Technology Major Project (2021SHZDZX0102), Shanghai
Qi Zhi Institute, Shanghai Science and Technology Commission (21511101200) and \href{https://openbayes.com/}{OpenBayes}. 
{\small
\bibliographystyle{ieee_fullname}
\bibliography{egbib}
}

\clearpage
\appendix

\section{VR-Folding Dataset}
\subsection{Task Definition}
In this section, we will describe the details of our task definition in our VR-Folding dataset. 
\subsubsection{Flattening}
\begin{figure*}[ht!]
  \centering
  \includegraphics[width=0.93\linewidth]{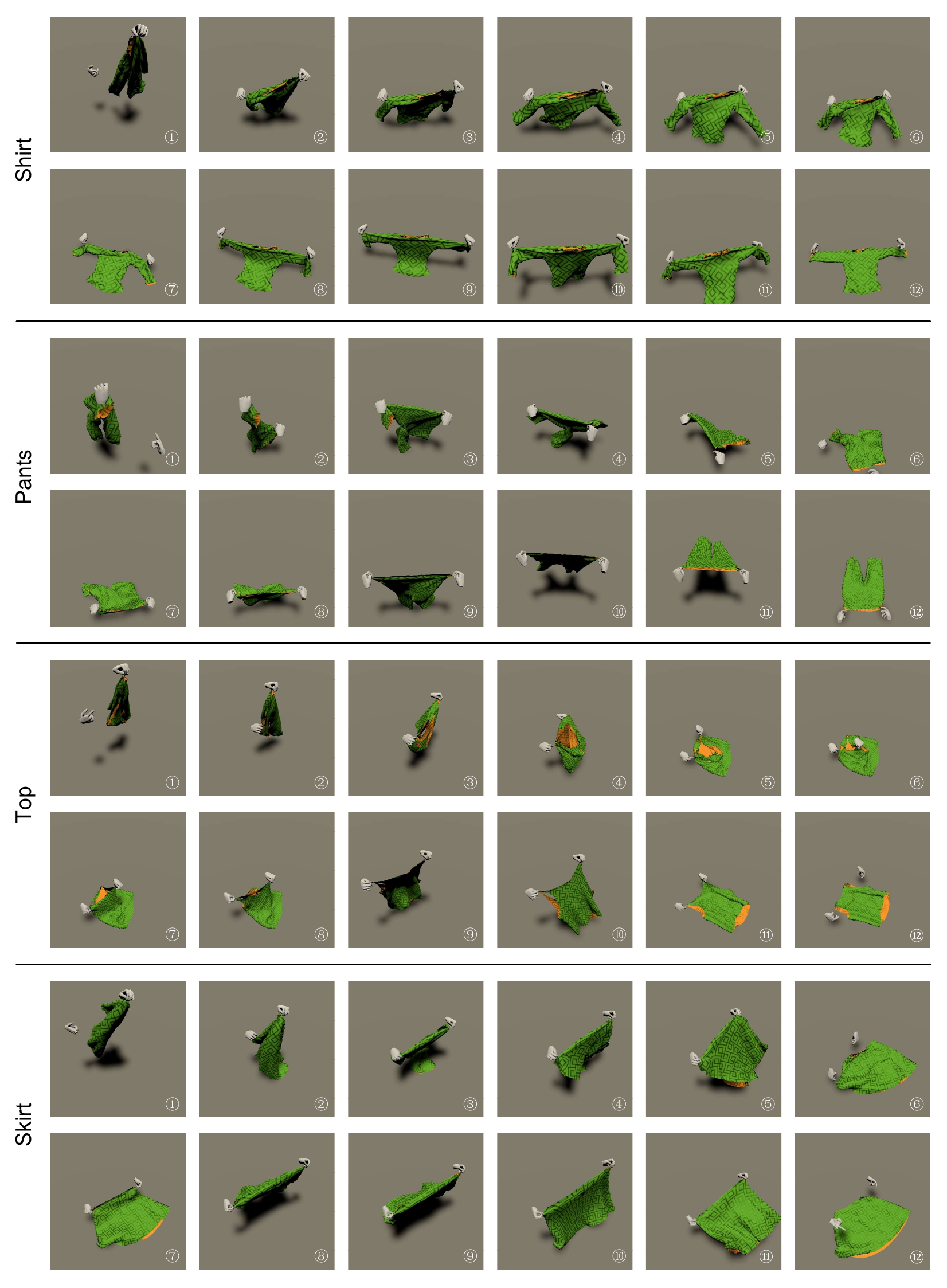}
   \caption{The examples of \textbf{Flattening} task.}
   \label{fig:flattening_example}
\end{figure*}

\cref{fig:flattening_example} gives some examples of \textit{Flattening} task. The goal of this task is making a garment in randomly crumpled state into a canonical flattened T-pose (see \cref{fig:flattening_example}) by a series of actions (\eg grasp, fling). Inspired by GarmentNets\cite{chi2021garmentnets} and FlingBot\cite{ha2022flingbot}, we firstly grasp the garment on one randomly selected point to simplify its initial configuration and increase its visibility, then fling the garment with both hands to flatten it efficiently. Specifically, this task can be divided into the following steps:
\begin{enumerate}
    \item \textbf{Grasp with a single hand}: The volunteer will grasp one randomly selected point on the garment with a single hand, and lift it in the air (see sub-figure [1] for Shirt in \cref{fig:flattening_example}). This initial configuration is similar to the definition in GarmentNets\cite{chi2021garmentnets}.
    \item \textbf{Grasp with both hands}: The volunteer will try to grasp two separate points on the garment with two hands simultaneously, and get ready for the following \textit{fling} action (see sub-figure [2] for Shirt in \cref{fig:flattening_example}).
    \item \textbf{Fling}: The volunteer will fling the garment with both hands to erase the wrinkles (see sub-figure [3-7] for Shirt in \cref{fig:flattening_example}).
    \item Repeat Step 2 and Step 3 until the garment is in flattened T-pose (see sub-figure [7-12] for Shirt in \cref{fig:flattening_example}). 
\end{enumerate}

\subsubsection{Folding}
\begin{figure*}[th!]
  \centering
  \includegraphics[width=0.93\linewidth]{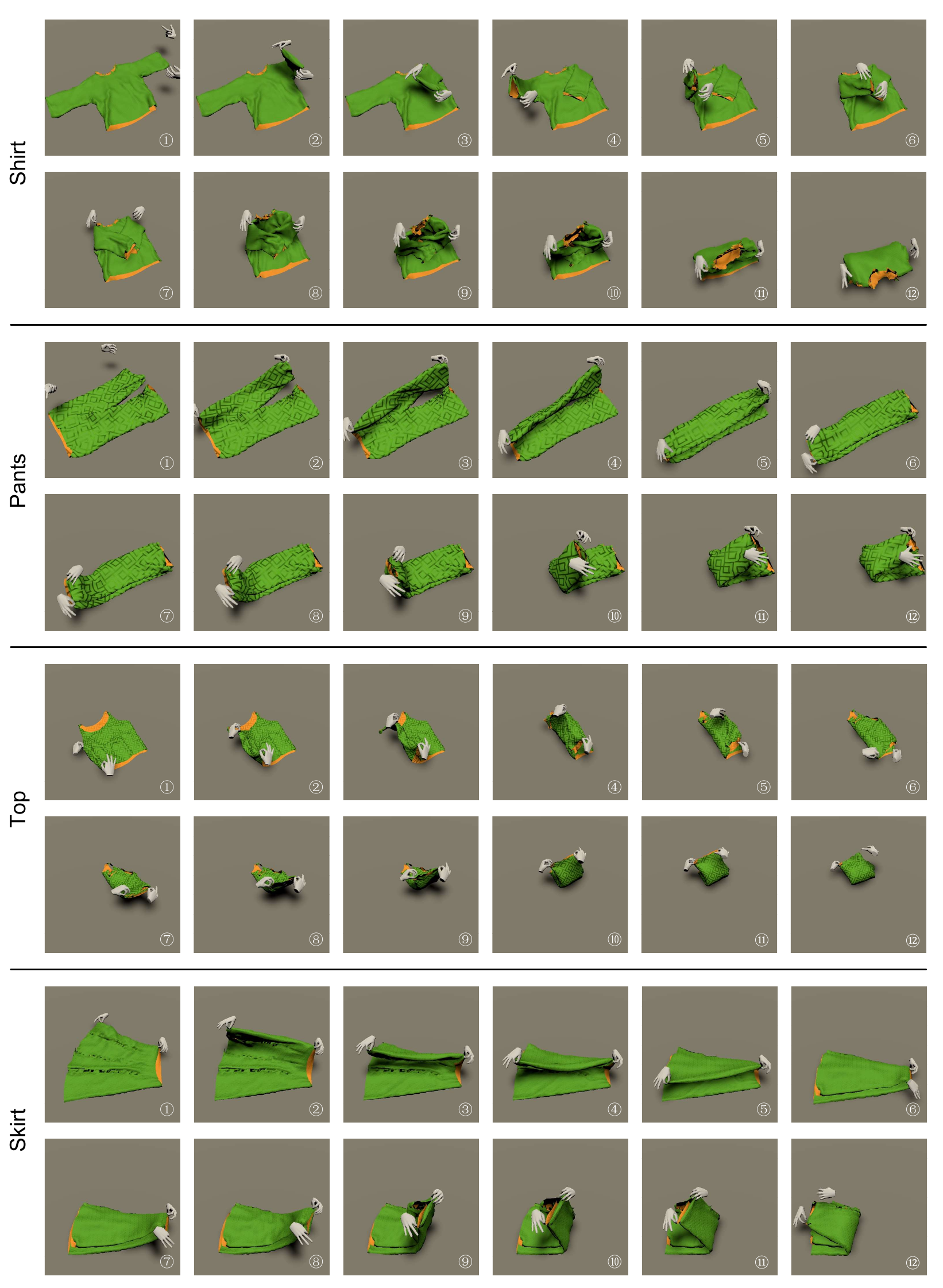}
   \caption{The examples of \textbf{Folding} task.}
   \label{fig:folding_example}
\end{figure*}

\cref{fig:folding_example} gives some examples of \textit{Folding} task. This task can be achieved by a series of bimanual pick-and-place actions. We make some general rules for the whole folding process. In order to enhance data variety, we encourage volunteers to fold garments in different ways (\eg fold from the left part or fold from the right part) without violating these rules. 
Here is the detailed rules of \textit{Folding} for each category:
\begin{enumerate}
    \item \textbf{Shirt}: The volunteer performs 2 (short-sleeve) or 3 (long-sleeve) pick-and-place actions with both hands for \textit{Shirt} (see examples for \textit{Shirt} in \cref{fig:folding_example}). For long-sleeved shirts, the first two actions will make the two sleeves folded, and the last action will make the trunk part folded. For short-sleeved shirts, the first action will fold in half along the left and right direction, and the last action will fold in half along the up and down direction.
    \item \textbf{Pants}: The volunteer performs 2 pick-and-place actions with both hands for \textit{Pants} (see examples for \textit{Pants} in \cref{fig:folding_example}). The first action will fold in half along the left and right direction, and the last action will fold in half along the up and down direction.
    \item \textbf{Top}: The volunteer performs 2 pick-and-place actions with both hands for \textit{Top} (see examples for \textit{Top} in \cref{fig:folding_example}). The first action will fold in half along the left and right direction, and the last action will fold in half along the up and down direction.
    \item \textbf{Skirt}: The volunteer performs 2 pick-and-place actions with both hands for \textit{Skirt} (see examples for \textit{Skirt} in \cref{fig:folding_example}). The first action will fold in half along the left and right direction, and the last action will fold in half along the up and down direction.
\end{enumerate}

\subsection{Data Statistics}

\begin{table}[h!]
\resizebox{0.48\textwidth}{!}{
  \centering
  \begin{tabular}{c|c|c|c|c|c}
    \toprule
    & & Shirt & Pants & Top & Skirt\\
    \midrule
    \multirow{2}{*}{Folding} & \# of instances & 989 & 1538 & 903 & 461\\
    & \# of videos & 993 & 1551 & 889 & 463 \\
    & \# of frames & 66830 & 77479 & 37214 & 23427\\
    \midrule
    \multirow{2}{*}{Flattening} & \# of instances & 1037 & 1631 & 1040 & 467 \\
    & \# of videos & 2145 & 1720 & 1063 & 943 \\
    & \# of frames & 243460 & 136411 & 132580 & 73357\\ 
    \bottomrule
  \end{tabular}
}
  \caption{Statistics of VR-Folding dataset.}
  \label{tab:data_stat}
\end{table}
The detailed data statistics of VR-Folding dataset is shown in \cref{tab:data_stat}. In the re-rendering process, we capture frames in Unity at the speed of 10 FPS. Each video only contains one instance, but one instance could occur in multiple videos. Each frame contains 4-view RGB-D images along with mask and complete garment mesh with NOCS labels. The multi-view RGB-D frames will be filtered with masks and transformed into point cloud. We split the whole dataset into \textit{train}, \textit{val}, \textit{test} subset on ratio $[0.8, 0.1, 0.1]$ by instances, which means that all the instances in videos of \textit{val} or \textit{test} subset are \textbf{unseen} during training. 
\section{GarmentTracking Network}
\subsection{3D Feature Extractor}
The 3D feature extractor (sparse ResUNet3D) used in our GarmentTracking network is based on FCGF\cite{choy2019fully}. The sparse 3D convolution operation is implemented by MinkowskiEngine\cite{MinkowskiEngine}. In our design, the output feature channels of each 3D convolution layer and transpose 3D convolution layer are $[64, 64, 128, 256]$ and $[64, 64, 64, 128]$ respectively. The feature dimension of output per-point feature is 64. Besides, we remove the L2-normalized operation for the final feature, because the relation attention module described below will perform L2-normalization for the feature later. Other hyper-parameters of the network structure are the same as FCGF\cite{choy2019fully}.

\subsection{Relation Attention Module (RAM)}
The self-attention module and cross-attention module are both based on the same Transformer-like structure called \textit{Relation Attention Module (RAM)} proposed by PTTR\cite{zhou2022pttr}. RAM can be formulated as $\operatorname{Att}(\mathbf{Q}, \mathbf{K}, \mathbf{V})$, which firstly projects input feature vectors of "Query (Q)", "Key (K)" and "Value (V)" into a latent feature space and then estimates the attention matrix between "Query" and "Key". The attention matrix is then applied to the "Value" feature to obtain the final attention product. 
Specifically, the self-attention and cross-attention operation can be formulated as \cref{eq:att1}, \cref{eq:att2} and \cref{eq:att3}:
\begin{align}
     \overline{\mathbf{X}}_1=\operatorname{Att}\left(\mathbf{X}_1, \mathbf{X}_1, \mathbf{X}_1\right) \label{eq:att1} \\ \overline{\mathbf{X}}_2=\operatorname{Att}\left(\mathbf{X}_2, \mathbf{X}_2, \mathbf{X}_2\right) \label{eq:att2} \\
    \hat{\mathbf{X}}=\operatorname{Att}\left(\overline{\mathbf{X}}_2, \overline{\mathbf{X}}_1, \overline{\mathbf{X}}_1\right) \label{eq:att3}
\end{align}
where $\mathbf{X}_1$ and $\mathbf{X}_2$ stands for features of previous frame and current frame respectively. $\overline{\mathbf{X}}_1, \overline{\mathbf{X}}_2$ are the self-attention features and $\hat{\mathbf{X}}$ is the cross-attention fusion feature. 
It is worth noting that RAM is a lightweight transformer, which only brings small computing overhead for the whole pipeline. 
The layer number and the head number of RAM are both 1. The feature dimension in the middle layer of RAM is 64. The MLP channels in the final feature layer is $[64, 128, 128]$. The dimension of the final fusion per-point feature is 128.

\subsection{NOCS Refiner}
The detailed network structure of NOCS Refiner is shown in \cref{tab:nocs_refiner}. We use ReLU as  activation function and use batch normalization in all the MLPs. During inference, the \textit{PC Refiner} will refine the raw NOCS class logits for each frame. However, we find that the \textit{Mesh Refiner} only needs very few steps to achieve reasonable results during tracking. So we only enable the \textit{Mesh Refiner} for the first few frames during long-term tracking for inference, which could increase the stability of the final outputs. For \textit{Folding} task, we enable it only in the first frame. For \textit{Flattening} task, we enable it in the first 15 frames in the video. 
\begin{table}[h!]
  \centering
  \begin{tabular}{c|c}
    \toprule
    Name & Feature Dimensions of MLP \\
    \midrule
    PC-PointNet & [326, 256, 256, 1024] \\
    Mesh-PointNet & [3, 64, 128, 1024] \\
    Mesh Fusion MLP & [2112, 512, 512, 1024] \\
    Mesh Refine MLP & [1024, 512, 256, 6] \\
    PC Refine MLP & [2304, 1024, 512, 192] \\
    \bottomrule
  \end{tabular}
  \caption{The detailed network structure of NOCS Refiner.}
  \label{tab:nocs_refiner}
\end{table}

\subsection{Details of Training and Inference}
The batch size in training is 16 for all of our experiments. During training, we randomly select two continuous frames in the same video as input. During inference, we use the predicted per-point NOCS coordinate and the refined canonical mesh of the previous frame as the input of the current frame. We remove the static frames (\ie frames without garment movement) and only track the moving frames (\ie frames that contain garment movement) in videos during inference.

\section{Additional Experiment Results}
\subsection{Noise Parameters for Robustness Experiment}
In the robustness experiment (Sec. 5.3.3) of the main paper, we augment the point-cloud NOCS coordinates of the first frame with a global scaling factor $\mathbf{s}_{pc}$, a global offset $\mathbf{o}_{pc}$, and Gaussian noise standard deviation $\delta$. Besides, we also augment the canonical mesh with global scaling factor $\mathbf{s}_{mesh}$. The detailed setting of these noise parameters is shown in \cref{tab:noise_params}.
\begin{table}
  \centering
  \begin{tabular}{c|c|c|c|c}
    \toprule
    Noise Level & $\mathbf{s}_{pc}$ & $\mathbf{o}_{pc}$ & $\delta$ & $\mathbf{s}_{mesh}$\\
    \midrule
    1x & $[0.8, 1.2]^3$ & $[0, 0.1]^3$ & 0.05 & $[0.8, 1.2]^3$ \\
    2x & $[0.6, 1.4]^3$ & $[0, 0.2]^3$ & 0.10 & $[0.6, 1.4]^3$ \\
    3x & $[0.4, 1.6]^3$ & $[0, 0.3]^3$ & 0.15 & $[0.4, 1.6]^3$ \\
    \bottomrule
  \end{tabular}
  \caption{The noise parameters in the robustness experiment. $[a, b]^3$ indicates a 3-D vector (\ie x, y, z axis) in which each dimension is uniformly sampled from $[a, b]$.}
  \label{tab:noise_params}
\end{table}
\subsection{Qualitative Results on VR-Folding Dataset}
\begin{figure*}[th!]
  \centering
  \includegraphics[width=0.94\linewidth]{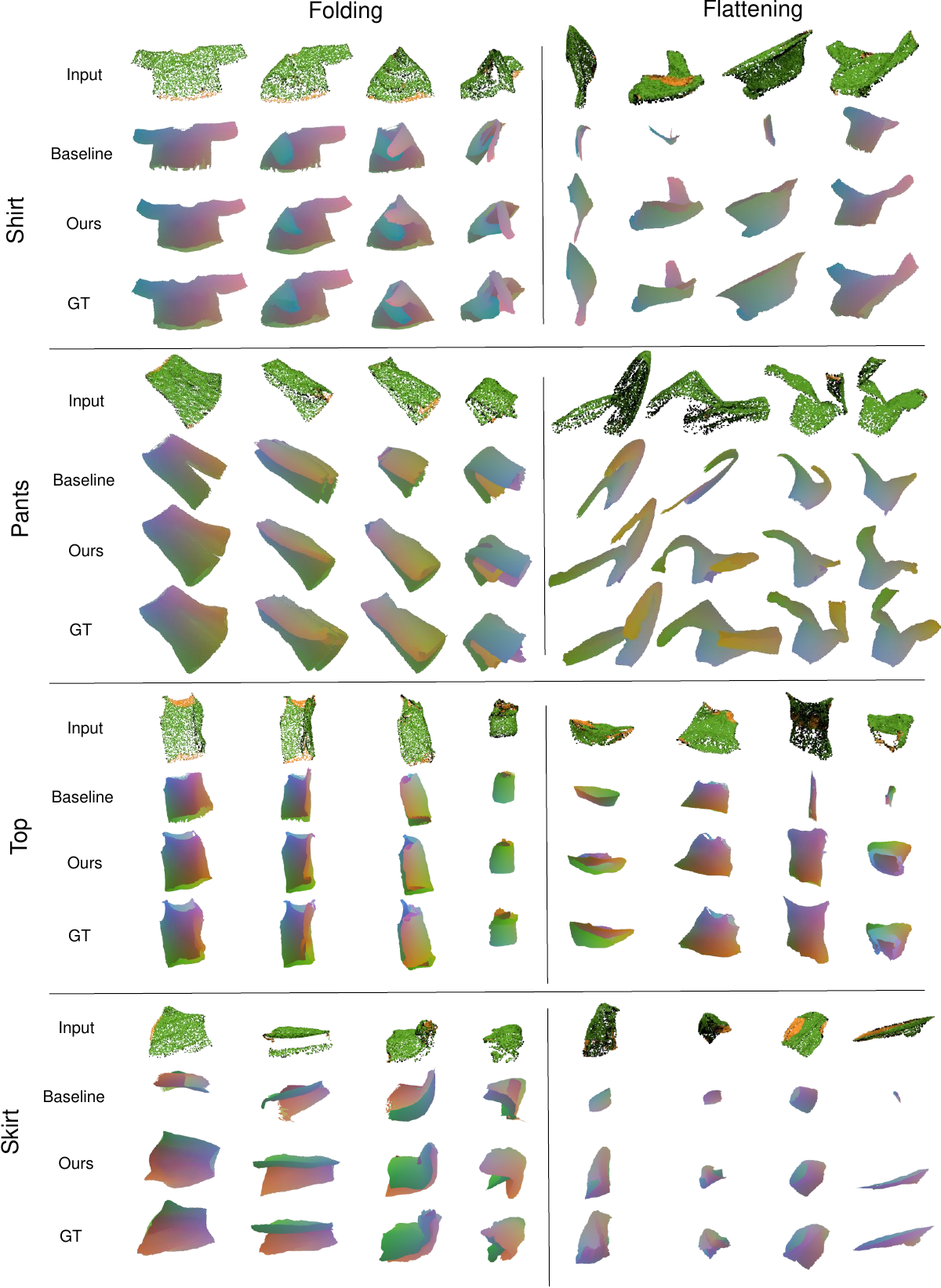}
   \caption{The additional qualitative results on \textbf{unseen} instances in VR-Folding dataset. \textbf{Please watch the video in the supplementary files for more elaborate examples}.}
   \label{fig:vis_sample_sim}
\end{figure*}

\cref{fig:vis_sample_sim} shows additional qualitative results on \textbf{unseen} instances in VR-Folding dataset. We use GarmentNets\cite{chi2021garmentnets} as the baseline here. \textbf{Please watch the video in the supplementary files for more elaborate examples}.

\subsection{Real-world Experiments}
We collect some real-world RGB-D ($640\times480$, 30FPS) videos of garment manipulation with 4 Realsense L515\cite{L515} LiDAR cameras. Before recording, We perform calibration (\ie intrinsic and extrinsic parameters) for the four L515 cameras. During recording, we ask the volunteer to repeat the same garment manipulation actions in VR-Folding dataset on \textbf{novel} garments in real world. After recording, we uniformly drop $3/4$ of the frames in the raw videos which makes the final videos all at 7.5FPS. Nextly, we perform post-processing for the recorded data. We firstly generate partial point cloud from the depth map of each camera then merge the multi-view partial point cloud into one single point cloud. Finally, we ask volunteers to segment the merged point cloud and only keep the garment part in the point cloud.

In order to narrow the gap between simulation and real world, we re-train our model and GarmentNets with pure depth information (\ie no RGB in point cloud) and perform zero-center operation for the input point cloud during training. During inference, we directly use the GarmentNets prediction (\ie canonical coordinates and mesh) as the first-frame pose. 

\begin{figure*}[th!]
  \centering
  \includegraphics[width=0.94\linewidth]{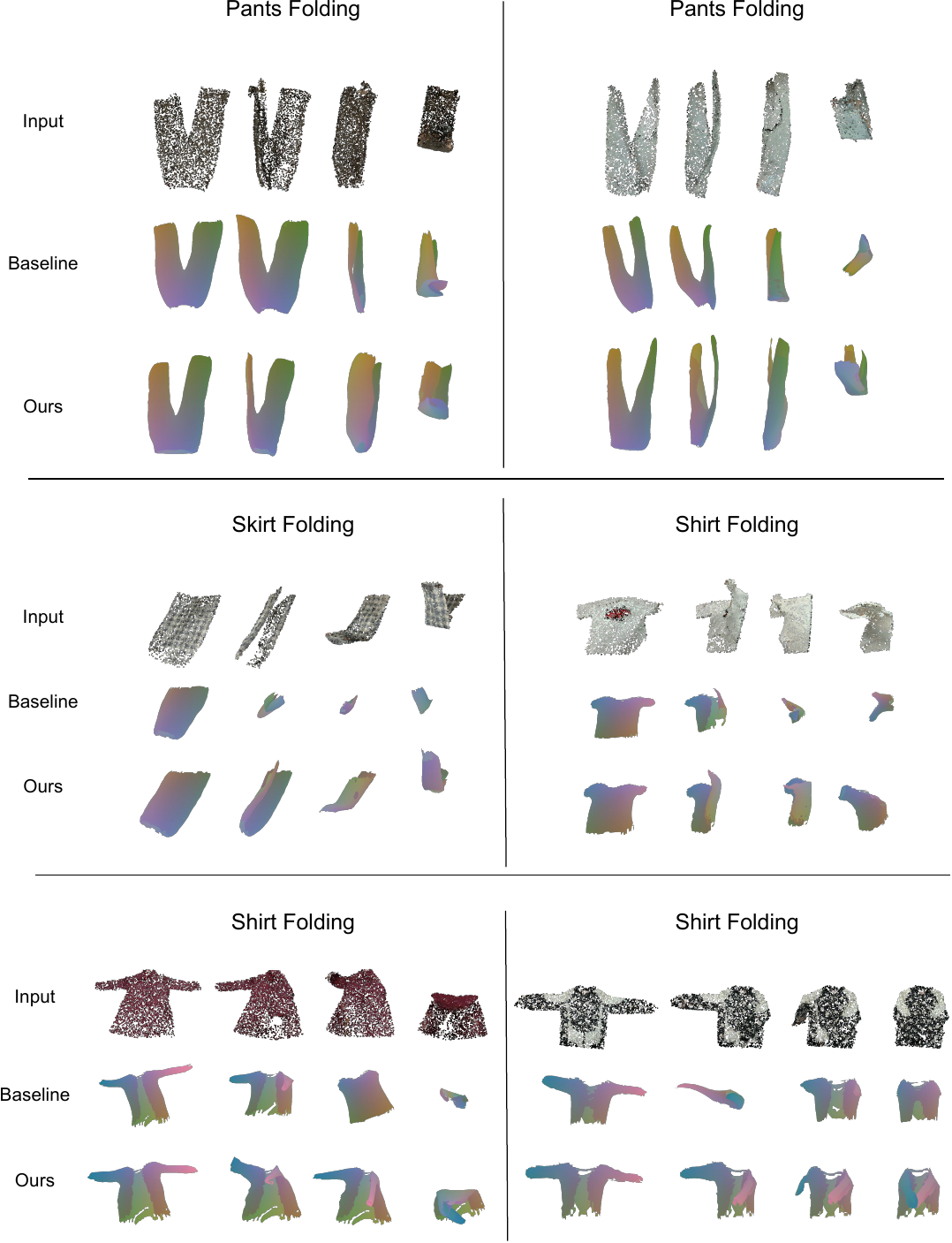}
   \caption{The qualitative results on \textbf{unseen} instances for \textit{Folding} task in \textbf{real-world} data. \textbf{Please watch the video in the supplementary files for more elaborate examples}.}
   \label{fig:vis_sample_real}
\end{figure*}

\begin{figure*}[th!]
  \centering
  \includegraphics[width=0.94\linewidth]{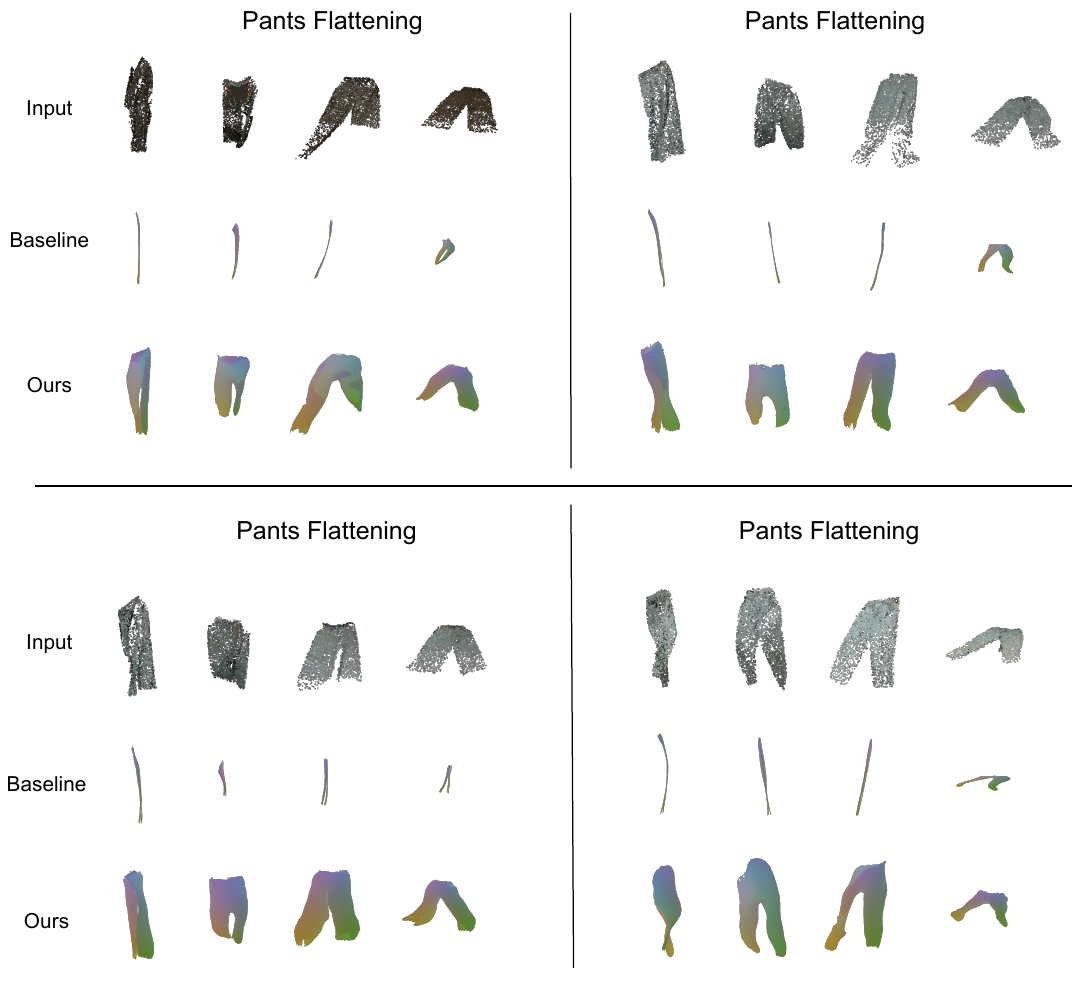}
   \caption{The qualitative results on \textbf{unseen} instances for \textit{Flattening} task in \textbf{real-world} data. \textbf{Please watch the video in the supplementary files for more elaborate examples}.}
   \label{fig:vis_sample_real_2}
\end{figure*}

\cref{fig:vis_sample_real} and \cref{fig:vis_sample_real_2} show some qualitative results on \textbf{unseen} garments in our collected real-world data. Our method can directly track pose for novel garments in the real-world with a model trained only on our simulated data, and exhibit more stability compared to the baseline (\ie GarmentNets). 
\textbf{Please watch the video in the supplementary files for more elaborate examples}.
\section{Broader Impact}
\begin{itemize}
    \item Since our dataset only contains the tasks associated with the daily garments, our work will not facilitate injury to living beings directly.
    \item We paid great attention to protect our volunteers' privacy in the process of collecting data.
    \item We respect human rights in all parts of our work.
    \item Our work only focuses on two tasks about deformable garments, so it's impossible to use our work to develop or extend harmful forms of surveillance.
    \item Our work causes no damage to the environment, because the dataset we built and used is colleted with a VR system.
    \item The garments appear in our VR-Folding dataset are all generated by software, so our dataset is unlikely to deceive people in real life.
\end{itemize}

\end{document}